\documentclass[12pt]{article}

\usepackage{amssymb}
\usepackage{svg}
\usepackage{siunitx}
\usepackage{amsmath}
\usepackage{amsthm}
\usepackage{tabularray}
\UseTblrLibrary{amsmath}
\UseTblrLibrary{siunitx}
\UseTblrLibrary{diagbox}
\usepackage{rotating}
\usepackage{mathtools}
\usepackage{pstricks}    
\usepackage{authblk}

\theoremstyle{definition}
\newtheorem{example}{Example}[section]
\newtheorem{definition}{Definition}[section]

\begin{document}



\title{Isolation Forest in Novelty Detection Scenario}
\author[1]{Adam Ulrich}
\author[1]{Jan Krňávek} 
\author[1]{Roman Šenkeřík}
\author[1]{Zuzana Komínková Oplatková}
\author[1]{Radek Vala}
\affil[1]{Faculty of Applied Informatics, Tomas Bata University in Zlin}


\maketitle
\section{Introduction}
\label{sec:introduction}




Datamining is a vast topic where we use automation mechanisms to process large data in various formats. That could be binary data from numerous electromechanical sensors, numerical data serialized by some processing computer, or even nominal data stored in the database.
Datamining algorithms can be used to find specific patterns in data, which is a topic of a pattern mining subfield. Algorithms in this subfield solve tasks like sequential mining of patterns \cite{agrawal1995mining} or frequent-itemset mining \cite{agrawal1994fast}. 
Some solutions lay in mining for similarities in data. Such similarities often form a batch in specific parts of analyzed space. It can be formed when specific attributes correlate. Another subfield of data mining focuses on identifying these batches --- also called clusters in data. Various clustering algorithms have been developed, such as DBSCAN \cite{Ester1996dbscan} or \(k\)-means \cite{lloyd1982kmeans} and their successful derivates.
With the recent upsurge of IOT, a subfield of anomaly data mining has become popular. Often, the data obtained is not what we expect it to be; sometimes, it can differ significantly from the rest and be identified as anomalies.


Anomaly detection problems can be viewed in various ways depending on the specific domain. There are anomaly detectors based on statistics (Z-score or Grubbs's test \cite{grubbs1949sample}), clusters \cite{he2003discovering} and density-based methods like Isolation Forest \cite{liu2008isolation, liu2012isolation}.


Usually, anomaly detection is used to solve one of the following tasks (see Markou et al. in \cite{MARKOU20032481}).

    \paragraph{Outlier detection} refers to the task with all of the data available in advance, and the algorithm is to identify outlying anomalies. This process is unsupervised since no labels are available in advance for the input data.
    \paragraph{Novelty detection} refers to the task with the majority of regular data available in advance and no anomalies. The algorithm is to learn how the regular data looks and later identify anomalies (in this scenario called \emph{novelties}).

Novelty detection is a semi-supervised technique for detecting anomalies (\emph{novelties}) unavailable in the training set.
The first algorithm mentioned in novelty detection is the One-class SVM algorithm \cite{tax2004support}. Despite being primarily developed for classification, this algorithm is often referred to as one of the first novelty detection algorithms. It learns from the input data it surrounds; hence, it can identify data outside this boundary.
This algorithm is based on quadratic programming, although Zhou et al. \cite{ZHOU20022927} provide an enhancement based on linear programming.
Another novelty detection algorithm is called the Local Outlier Factor from the family of distance-based algorithms \cite{breunig2000lof} that assigns each object a degree of being an anomaly.

One of the downsides when dealing with the above algorithms is their lack of interpretability. It is not trivial to visualize the outcomes and understand the reasons for the point being a novelty or otherwise, making it arduous to understand the dataset's properties. That is not the case for the Isolation Forest, though, where the visualization is its superiority.
Elaborate research on novelty detectors has highlighted several determining properties when dealing with such problems. First, the novelty detection algorithm should be semi-supervised (that is, it can be trained on a dataset first and evaluated later). Then, the algorithm should be able to work with \(n\)-dimensional spaces. Lastly, the algorithm should evaluate both data seen during the learning phase and those never seen before. Only then will the novelty detector be able to work properly.

The Half-Space Tree (HST) algorithm, introduced by Tan et al. in \cite{tan2011fast} for anomaly detection for streaming data, is a data structure that recursively partitions the data space into half-spaces to model the distribution of normal instances.
The core of this algorithm is based on the Isolation Forest by Liu et al. \cite{liu2008isolation}, \cite{liu2012isolation}.
It builds on the idea of the binary decision tree and alters the process of building and evaluating the decision tree to isolate anomalies. By using an algorithm based on binary trees, the interpretability of the problem is significantly enhanced, as the tree structure allows us to visually trace and understand the decision-making process.

Our main task is to utilize the HST algorithm to develop a solid and adaptable novelty detector.
This article proposes a new HST modification, similar to Ting's in \cite{ting2013mass}, specifically tailored for detecting novelties.
We show that this modification is particularly well-suited for novelty detection due to its unique structure and operational approach.
First, we provide a theoretical framework for this modified algorithm and the original Isolation Forest.
We then provide examples of using it to build and evaluate the tree.
With this framework built up, we show the distinction between those two algorithms and examples of using them in novelty search scenarios.

\section{Glossary}
\label{sec:theory}

Since this article is centered around anomaly detection algorithms, several key terms used throughout the article are introduced below.

\begin{itemize}
    \item \emph{Data point} refers to any observable data with \(n\) dimensions.
\item Regular point is a data point included in the given dataset. Its features are expected.
\item \emph{Anomaly} is a data point that differs significantly from other observations.
\item \emph{Outlier} is an anomaly included in the given dataset.
\item \emph{Novelty} is an anomaly that is not present in the given dataset during learning. Novelties are supplied later during evaluation.
\item \emph{Supervised} algorithm is an algorithm that is trained on all available data labels.
\item \emph{Unsupervised} algorithm is an algorithm that does not get any labels for the training data.
\item \emph{Semi-supervised} algorithm is an algorithm that is trained on data of only one class.

\end{itemize}

\section{Graph Theory}
\label{sec:graph_theory}

 The isolation forest is composed of trees that are, in fact, directed graphs.
 Hence, the theoretical framework used in this article is centred mainly around graph theory.
 The definitions used in this section are based on Rosen et~al. \cite{rosen2012discrete}.

\begin{definition}
 \emph{Directed graph} (or digraph) $D = (V, E)$ consists of a nonempty set of vertices $V$ and a set of directed edges $E \in V \times V$.
 Each directed edge is an ordered pair of vertices.
 The directed edge $(u, v)$ is said to start at $u$ and end at $v$.
\end{definition}

For the digraph $D = (V,E)$ we define:
\begin{itemize}
    \item \emph{Descendant} of a vertex $v \in V$ is a vertex $v' \in V$ such that $(v,v') \in E$. 
    We say that $v$ is the \emph{ancestor} associated with $v'$.
    \item \emph{Walk} $W$ from a vertex $v_0 \in V$ to a vertex $v_n \in V$ is a finite set of edges ($W \subseteq E$), such that $$W = \{(v_0, v_1),(v_1, v_2),\dots,(v_{n-1}, v_n)\}.$$
    The empty set $W = \emptyset$ is a trivial walk from $v$ to $v$ for every $v \in V$. 
\end{itemize}

With the isolation tree starting at the root, we define a rooted tree.

\begin{definition}
\emph{Rooted} tree is the digraph $T = (V,E)$ with a single vertex $r \in V$ designated as the \emph{root}, with exactly one walk (so-call \emph{path}) from $r$ to $v$ for each $v \in V$.
\end{definition}

For the rooted tree $T = (V,E)$ we define:
\begin{itemize}
    \item \emph{Depth} of $v \in V$ is the size of the path from the root to $v$.
    \item \emph{Leaf} is a vertex $v \in V$ that has no descendants. 
    \item \emph{Internal  vertex} conversely has a descendant.
    \item To \emph{traverse} a tree $T$ is to take a path $P$ from the root to the leaf. If $(v,v') \in P$, we say that we \emph{visit} $v$ and $v'$ while traversing.
\end{itemize}

\section{Isolation Forest}
\label{sec:isolation_forest}

Isolation Forest
\cite{liu2008isolation, liu2012isolation} is an outlier
detection, unsupervised ensemble algorithm. This approach is well-known for successfully identifying outliers using recursive partitioning (forming a decision tree) to decide whether the analyzed data point is an anomaly. The fewer partitions required to isolate, the more probable it is for a point to be an anomaly.

The initial problem with using the original Isolation Forest for novelty detection is that a potential novelty point located far from training data tends to fall into some existing branch because the Isolation Forest's splits were created without seeing it, making its placement arbitrary and leading to incorrect isolation.
This statement will later be supported by mathematical analysis and revisited in \ref{sec:revisiting}.

In the following sections, we describe the original decision tree.
We deviated slightly from Liu et al.~\cite{liu2008isolation} for better interpretability and comparability with the proposed solution.
Note that this deviation is theoretical and does not affect the original functionality.

\subsection{Definitions}
\begin{definition}
Let $\mathsf{s}=(s_0, \dots, s_d, \dots, s_{n-1})\in \mathbb{R}^n$ be a data point. Then we say that $\pi_d(\mathsf{s}):=s_d$ is a \emph{projection} of data point $\mathsf{s}$ onto  dimension $d$  yielding $s_d$.
\end{definition}

\begin{definition}
Let $Z$ be a finite subset of $\mathbb{R}^n$,
$$Z \subseteq \mathbb{R}^n ;\quad n \in \mathbb{N}.$$

For each dimension \(d \in\{0, \dots, n - 1\}\), let
$$Z_d = \{ \pi_d(\mathsf{s})\ |\ \mathsf{s} \in Z \}.$$

Then we define a hyperrectangle $R(Z)$ \emph{surrounding} $Z$, such that
$$R(Z) = r_0 \times r_1 \times \cdots \times r_{n-1},$$ where $r_d = \langle \min Z_d, \max Z_d \rangle$ for each $d \in \{0,1, \dots, n-1\}.$

\end{definition}

\subsection{Constructing a decision tree}
\label{sec:cdt}
\begin{enumerate}

    \item Each leaf or internal vertex is a hyperrectangle $R$. 
    \item Function relation $\rho$ assigns the split point $z$ and the dimension $d$ to the internal vertex $v$,
    $$(v, (z,d)) \in \rho.$$
    \item The sample \(S\) contains \emph{b} number of input points of $n$ dimensions, that is
    $$S \subseteq \mathbb{R}^n ;\quad |S| = b; \quad b \in \mathbb{N}.$$
    \item The ending condition of the leaves is satisfied if the vertex \(R\) satisfies \[| S \cap R | = 1.\]
\end{enumerate}

\paragraph{Basis step}
The trivial rooted tree \(T_0\) is a tuple with
vertices \(V_0 = \{R(S)\}\) and edges \(E_0 = \emptyset\), i.e. 
\[T_0= (V_0, E_0) = (\{R(S)\},\emptyset).\]
The function relation $\rho_0$ initially has no assignments, i.e.
$$\rho_0 = \emptyset.$$

\paragraph{Recursive step}
The task of this step is to reach the tree \(T_{j+1}\) from \(T_{j} = (V_j, E_j)\) and $\rho_{j+1}$ from $\rho_{j}.$
Let \(L_j \subseteq V_j\) be a subset of leaves not satisfying the
ending condition
and \(R \in L_j\)\ be a leaf such that
\[R =  r_0 \times \cdots \times r_{d_R} \times \cdots \times r_{n-1}. \]
We first select a random dimension $d_R$ (such that $r_{d_R}$ is infinite) and a random split point $z_R \in r_{d_R}$.
The split point $z_R$ splits $R \cap S$, creating two disjunctive sets $S_l$ and $S_r$ respectively, such that
\begin{align*}
S_l &= \{ \mathsf{s} \in{R \cap S}\ |\ \pi_{d_R}(\mathsf{s})\le z_R\},\\
S_r &= \{ \mathsf{s} \in{R \cap S}\ |\ \pi_{d_R}(\mathsf{s}) > z_R\}.
\end{align*}
Then we obtain left and right hyperrectangles \(R_l\), \(R_r\) as
follows:
\begin{align*}
R_l &= R(S_l),&
R_r &= R(S_r).
\end{align*}

Each vertex \(R \in L_j\) is associated with two new
edges \((R,R_l ), (R, R_r)\) and is assigned with $(z_R,d_R)$ by function relation $\rho_{j+1}$, such that

\begin{align*}
   \rho_{j+1} &= \rho_j \cup \bigcup_{R \in L_j} \{(R, (z_R, d_R))\}, \\
   V_{j+1} &= V_j \cup \bigcup_{R \in L_j} \{R_l, R_r\}, \\
   E_{j+1} &= E_j \cup \bigcup_{R \in L_j} \{(R, R_l), (R,R_r)\},\\
   T_{j+1} &= (V_{j+1}, E_{j+1}),
\end{align*}
i.e.~${R_l, R_r} \subset R$ are leaves and $R$ is an inner vertex in the new tree
\(T_{j+1}\).\footnote{Tree \(T_{j+1}\) is actually a Hasse diagram of the ordered set
\((V_{j+1},\subseteq)\)}

\paragraph{Termination} The algorithm moves to the next recursion step unless there is an equality of two consecutive trees \(T_k = T_{k+1}\). Such equality happens when all leaves satisfy their ending condition, i.e., \(L_k = \emptyset\).
Thus, the desired tree $T$ is the tree $T_k$, and the finite relation $\rho$ is $\rho_k$.

\subsection{Example of decision tree construction}
\label{example:original_tree_create}
\begin{figure*}[!t]
\centering
\includegraphics[width=0.9\textwidth]{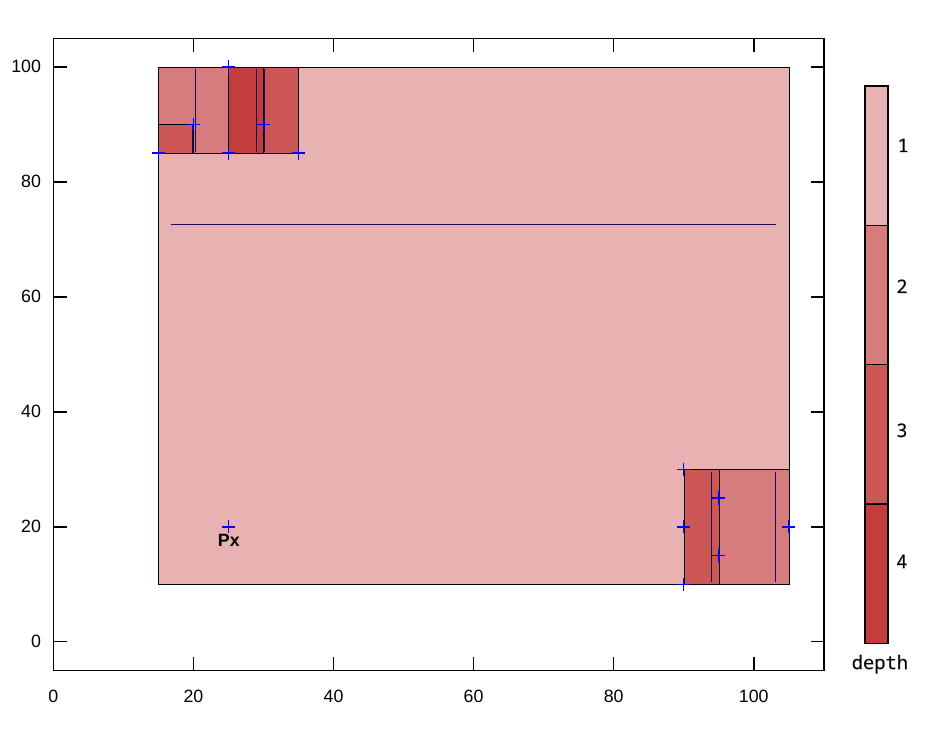}
\caption{Original solution. Rectangles created by recursive splitting.}
\label{fig:example_noutlier_gnu}
\end{figure*}

Consider now an example of creating a new Isolation tree based on the given input sample
\begin{align*}
    S = \left\{\begin{smallmatrix}
    [25,100],&[30,90],&[20,90],&[35,85],\\
    [25,85],&[15,85],&[105,20],&[95,25], \\
    [95,15],&[90,30],&[90,20],&[90,10]
    \end{smallmatrix}\right\},
\end{align*}
as shown in Figure \ref{fig:example_noutlier_gnu}.

The tree $T$ is created by starting with the tree $T_0$ and expanding further as described by the recursive step until the ending condition is met.
Figure \ref{fig:example_noutlier_tree_color} shows the finished tree $T$, trained on the dataset $S$. Numbers represent the final leaves’ depth.

\paragraph{Basis step} 
We create the tree $T_0$ and a function relation $\rho_0$.
   There is just a single vertex (root) 
   \[R = R(S) = \langle 15, 105 \rangle \times \langle 10, 100 \rangle,\]
   without any connections, so both $E_0$ and $\rho_0$ are empty, i.e.
\begin{align*}
V_0 &= \{R\},& E_0 &= \emptyset,\\
T_0 &= (V_0, E_0),& \rho_0 &= \emptyset.
\end{align*}
Figure \ref{fig:example_noutlier_gnu} shows $R$ as the largest and the brightest coloured rectangle.

\paragraph{Recursive step} 
In order to reach $T_1$ from $T_0$, a dimension $d_R=1$ and a split point $z_R = 72.63$ were randomly chosen. Figure \ref{fig:example_noutlier_gnu} shows split points as purple lines in their respective rectangles. The recursive step starts with creating left and right descendants of $R$ as
\begin{align*}
S_l &= \left\{\begin{smallmatrix}
 [105,20],& [95,25], &[95,15],\\
[90,30],&[90,20],&[90,10]
\end{smallmatrix}\right\}, \\
S_r &= \left\{\begin{smallmatrix}
    [25,100],&[30,90],&[20,90],\\
    [35,85], &[25,85],&[15,85]
\end{smallmatrix}\right\}.
\end{align*}
forming a new tree $T_1$ and a new function relation $\rho_1$ as follows:
\begin{align*}
V_1 &= \{R, R_l, R_r\},&
E_1 &= \{(R,R_l), (R,R_r)\},\\
T_1 &= (V_1, E_1),&
\rho_1 &= \{ (R, (72.63, 1))\}.
\end{align*}
The tree $T_1$ has the left leaf $R_l$ and the right leaf $R_r$; the ending condition is not met, that is $L_1 = \{R_l,R_r\}$. 

We continue the recursive step with two vertices.
For the left vertex $R_l$, the dimension $d_{R_l} =0$ and the split point $z_{R_L}= 103.08$ were chosen randomly, giving
\begin{align*}
S_{ll} &= \left\{\begin{smallmatrix}
    [95,25],& [95,15],\\ 
    [90,30],&[90,20],\\
    [90,10]
\end{smallmatrix}\right\},&
S_{lr} &= \{[105,20]\},\\
R_{ll} &= \langle 90, 95 \rangle \times \langle 10, 30\rangle,&
R_{lr} &= \langle 105, 105 \rangle \times \langle 20, 20\rangle
\end{align*}
and $z_{R_r}= 20.32$, $d_{R_r} = 0$ for the right vertex $R_r$ respectively
\begin{align*}
S_{rl}&= \{[20,90],[15,85]\},&
S_{rr} &= \left\{\begin{smallmatrix}
    [25, 100],& [30,90],\\
    [35,85],& [25,85],
\end{smallmatrix}\right\},\\
R_{rl}&= \langle 15, 20 \rangle \times \langle 85, 90 \rangle,&
R_{rr}&= \langle 25, 35 \rangle \times \langle 85, 100 \rangle.
\end{align*}

With the rectangles prepared, we can assemble new vertices and edges and create a new $T_2$:
\begin{align*}
\rho_2 &= \{(R,(72.63,1), (R_l, (103.08, 0)), (R_r, (20.32, 0)) \},\\
V_2 &= \{ R, R_l, R_r, R_{lr}, R_{lr}, R_{rl}, R_{rr} \},\\
E_2 &= \left\{\begin{smallmatrix} 
(R,R_l),&(R,R_r), &(R_l,R_{ll}),\\ 
(R_l,R_{lr}),& (R_r,R_{rl}),& (R_r,R_{rr})
\end{smallmatrix}\right\},\\
T_2 &= (V_2, E_2).
\end{align*}

\paragraph{Termination} With the $T_2$ created, we now have to check for the ending condition of the leaves. Since $|R_{lr}\cap S|=|S_{lr}| = 1$, the ending condition for the leaf $R_{lr}$ is met, and the new set of leaves $L_2$ for the next recursion step is
$$L_2 = \{R_{ll},R_{rl},R_{rr}\}.$$

This continues until we reach tree $T_5$ such that $T_5=T_6$ is the desired tree $T$ as shown in Figure \ref{fig:example_noutlier_tree_color}.

\subsection{Evaluating decision tree}
The evaluation of desired element $a$ starts in the root $R$ of the previously built tree $T$.
In a root, by applying function relation, $\rho(R) = (z,d)$, we obtain split point $z$ and dimension $d$.
The root $R$ of a tree $T$ has two descendants $R_l$, $R_r$, such that
$\forall r_l\in R_l; \pi_d(r_l) \le z$ and $\forall r_r\in R_r; \pi_d(r_r)  > z$.

\begin{figure*}[!t]
\includegraphics[width=1\textwidth]{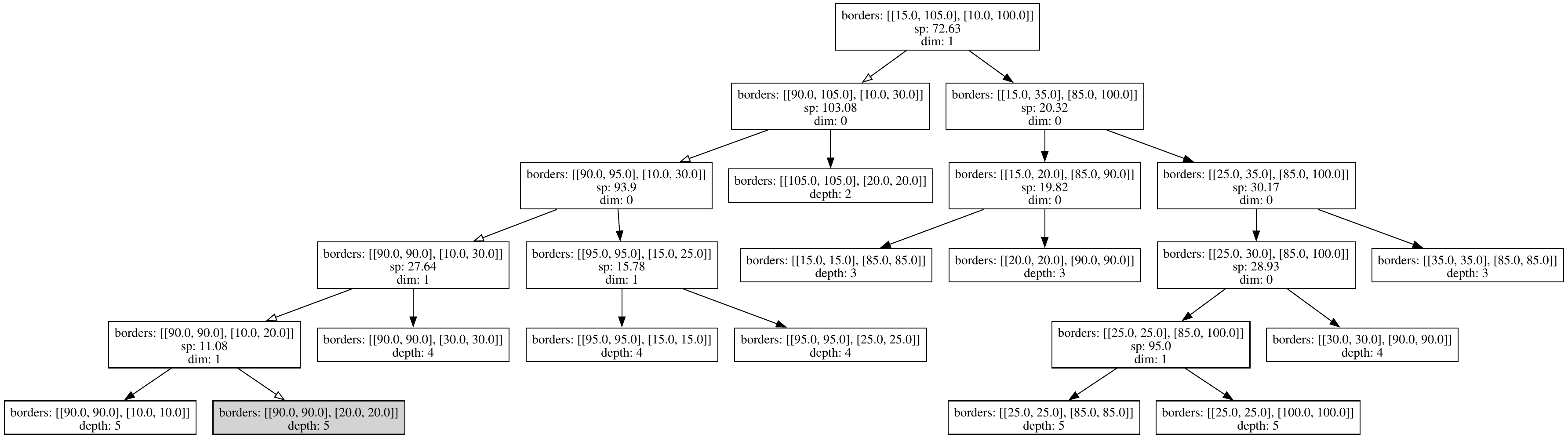}
\caption{tree constructed using the original approach}
\label{fig:example_noutlier_tree_color}
\end{figure*}

If $\pi_d(a)\le z$, we visit $R_l$, else, that is $\pi_d(a) > z$, we visit $R_r$.
We continue in this manner until we reach the leaf. Figures \ref{fig:example_noutlier_gnu} and \ref{fig:example_noutlier_tree_color} show the final depths of individual leaves. Based on them, we can decide on the level of outlierness. The deeper the evaluated point in the tree, the less anomalous it gets.

\begin{example}
\label{ex:regular_point_evaluation_original}
    Consider evaluating the point $a = [105,20]$ on the tree $T$ built in Example \ref{example:original_tree_create}.

    We start with the root $R = \langle 15,105\rangle \times \langle 10, 100 \rangle$.
    By applying the function $\rho(R)$, we obtain the split point $z = 72.63$ and the dimension $d = 1$.
    The root $R$ has two descendants 
\begin{align*}
    &R_l = \langle 90,105\rangle \times \langle 10, 30 \rangle,&
    &R_r = \langle 15,35\rangle \times \langle 85, 100 \rangle,\\
    \intertext{such that}
    &\forall r_l\in R_l; \pi_1(r_l) \le 72.63,&
    &\forall r_r\in R_r; \pi_1(r_r) > 72.63.
\end{align*}
Now, by applying the projection $\pi_1$ on $a$, we obtain $20$, which is less than the split point $z = 72.63$, i.e.
$$\pi_1([105,20]) = 20 < 72.63.$$
We visit the vertex $R_l$ because the value obtained by applying the projection on any element of $R_l$ is smaller than $72.63$.

We reached the next recursive step. With the vertex $R_l$ visited, we apply the function $\rho(R_l)$, obtaining the new split point $z = 103.08$ and the new dimension $d = 0$.
The vertex $R_l$ has two descendants 
\begin{align*}
    &R_{ll} = \langle 90,95\rangle \times \langle 10, 30 \rangle,&
    &R_{lr} = \langle 105,105\rangle \times \langle 20,20 \rangle,\\
    \intertext{such that}
    &\forall r_{ll}\in R_{ll}; \pi_0(r_{ll}) \le 103.08,&
    &\forall r_{lr}\in R_{lr}; \pi_0(r_{lr}) > 103.08.
\end{align*}

We visit the vertex $R_{lr}$ because the value obtained by applying the projection $\pi_0$ on any element of $R_{lr}$ and also on $a$ is more than $103.08$.

Since $R_{lr}$ has no descendants, we reached the final leaf with the depth of~$2$.
\end{example}

\begin{example}
\label{ex:novelty_point_evaluation_original}
    Consider now the evaluation of $a' = [25,20]$ on the tree $T$ built in Example \ref{example:original_tree_create}. Note that $a'$ was not contained in the training set for building a tree.

    We start with the root $R = \langle 5,105\rangle \times \langle 10, 100 \rangle$.
    By applying function $\rho(R)$, we obtain the split point $z = 72.63$ and the dimension $d = 1$.
We visit $R_l = \langle 90,105\rangle \times \langle 10, 30 \rangle$ 
because $\pi_1([25,20]) = 20 \le 72.63.$

We reached the next recursive step. Obtaining the new split point $z = 103.08$ and the new dimension $d = 0$, we visit $R_{ll} = \langle 90,95\rangle \times \langle 10, 30 \rangle$, since $\pi_0([25,20]) \le 103.08$.


This repeats recursively until the leaf $R_{llllr}$ is reached. Note that this is the leaf with the point $[90,20]$. The reached depth is 5, as shown in Figure \ref{fig:example_noutlier_tree_color} (the leaf is marked grey).

Figure \ref{fig:example_noutlier_gnu} shows that after just two steps, the $[25,20]$ is no longer a part of any further evaluated rectangles.

\end{example}

Note that each element that was part of the batch during the training --- tree building --- phase is always contained in each vertex it visits. See $a \in R_{lr} \subset R_{l} \subset R$ in Example \ref{ex:regular_point_evaluation_original}.
This is not true for elements unseen during the training phase (such as novelty points).
See $a' \in R$, but $a' \notin R_l$ (and of course $a' \notin R_r$) in Example \ref{ex:novelty_point_evaluation_original}.

\subsection{Revisiting the Initial Problem}
\label{sec:revisiting}
Suppose we now want to use the original tree to evaluate a novelty data point $p$, which is not present in the training set.
Let us recall the equations from \textit{recursive step} in section \ref{sec:cdt} needed for vertex creation:
\begin{align*}
S_l &= \{ \mathsf{s} \in{R \cap S}\ |\ \pi_{d_R}(\mathsf{s})\le z_R\},\\
S_r &= \{ \mathsf{s} \in{R \cap S}\ |\ \pi_{d_R}(\mathsf{s}) > z_R\}.
\end{align*}
The point $p$ fits neither $S_l$ nor $S_r$ and does not even necessarily fit $R_l = R(S_l)$ nor $R_r = R(S_r)$.\footnote{For practical illustration, see point $a'$ in Example \ref{ex:novelty_point_evaluation_original}.}

Nevertheless, because the point $p$ satisfies  $\pi_d(p) \le z$ (or $\pi_d(p) > z$), $p$ \emph{is assigned} to the vertex $R_l$ (or $R_r$) even though $p \notin R_l$ nor $p \notin R_r$.


\section{Proposed Half-Space Tree novelty detector}
\label{sec:novelty_isolation_forest}
This section proposes a new novelty detector based on the Half-Space tree algorithm.
This modification takes the basic idea of an ensemble of trees with various depths but takes it further to make semi-supervised detection possible.

\subsection{The Solution}
The proposed solution comes from the idea that the original tree lacks the possibility to isolate more data points than it currently observes.
The observed space is bounded by the minimum and maximum in each dimension.

As in the original article, we use the concept of a binary decision tree. The proposed solution is altering the idea of the split point evaluation. Whereas the original Isolation Forest evaluates the split point based on the previous data, the HST algorithm evaluates the split point based on a range. For this to work, several alterations to the split point evaluation and the form of data passed between vertices must be made; however, the overall concept of the forest remains the same.
The HST algorithm has several key concepts, which we adopted and modified as follows:

\begin{enumerate}
    \item We start with the root, representing the whole domain space bounded by ranges.
    \item Descendants cover the whole observable space of their associated ancestors. 
    \item The split point is in the \emph{middle} of the given dimension’s range.
    \item The input data is only used to determine the ending condition.
\end{enumerate}

\subsection{Constructing the decision tree}

\begin{enumerate}
    \item The sample \(S\) is the nonempty set of input points.
    \item Leaves and internal vertices are possibility-space hyperrectangles. 
    \item A hyperrectangle $R$ satisfies the ending condition when \(S \cap R = \emptyset\) or \(| S \cap R | = 1\).
\end{enumerate}

\paragraph{Basis step}
Each dimension \(d \in\{0, \dots, n-1\}\), is bounded by the range \(r_d\). The ranges form the possibility-space hyperrectangle \(R_0\), i.e.
\[R_0 =  r_0 \times r_1 \times \cdots \times r_{n-1}.\]

The trivial rooted tree \(T_0\) is a tuple with
vertices \(V_0 = \{R_0\}\) and edges \(E_0 = \emptyset\), i.e.
\[T_0= (V_0, E_0) = (\{R_0\},\emptyset).\]

\paragraph{Recursive step}
The steps to reach the tree \(T_{j+1}\) from \(T_{j} = (V_j, E_j)\) are
as follows:

Let \(L_j \subseteq V_j\) be a subset of leaves not satisfying the
ending condition.
For each leaf \(R \in L_j\) we
select a random dimension \(d\).

Let
\begin{align*}
R &= r_0 \times  \cdots \times r_{d-1} \times  r_d\times r_{d+1} \times \cdots \times r_{n-1},
\end{align*}
where $r_d = \langle x, y )$.
Then, we obtain the left and right hyperrectangles
\begin{align*}
R_l &= r_1 \times  \cdots \times r_{d-1} \times  \langle x, s ) \times r_{d+1} \times \cdots \times r_n, \\
R_r &= r_1 \times  \cdots \times r_{d-1} \times \langle s, y ) \times r_{d+1} \times \cdots \times r_n,
\end{align*}
where \(s = \frac{x + y}{2}\) is a number obtained as the middle of the range \(r_d\,\).

In the new tree $T_{j+1}$ each vertex \(R\) is associated with two new
edges \((R,R_l ), (R, R_r)\) giving
\begin{align*}
V_{j+1} &= V_j \cup \bigcup_{R \in L_j} \{R_l, R_r\},\\
E_{j+1} &= E_j \cup \bigcup_{R \in L_j} \{(R, R_l), (R,R_r)\},\\
T_{j+1} &= (V_{j+1}, E_{j+1}),
\end{align*}
i.e.~${R_l, R_r} \subset R$ are leaves in the new tree
\(T_{j+1}\).

\paragraph{Termination} Recursion is terminated if there is an equality of two consecutive trees \(T_k = T_{k+1}\). This happens when all leaves satisfy the ending condition, i.e., \(L_k = \emptyset\).
If this is the case, the desired tree $T$ is the tree $T_k$; otherwise, move to the next recursion step.

Note that the tree \(T_{j}\) is actually a Hasse diagram of the ordered set
\((V_j,\subseteq)\).

\subsection{Example of decision tree construction}
\label{example:novelty_tree_create}
\begin{figure*}[!t]
\centering
\includegraphics[width=0.9\textwidth]{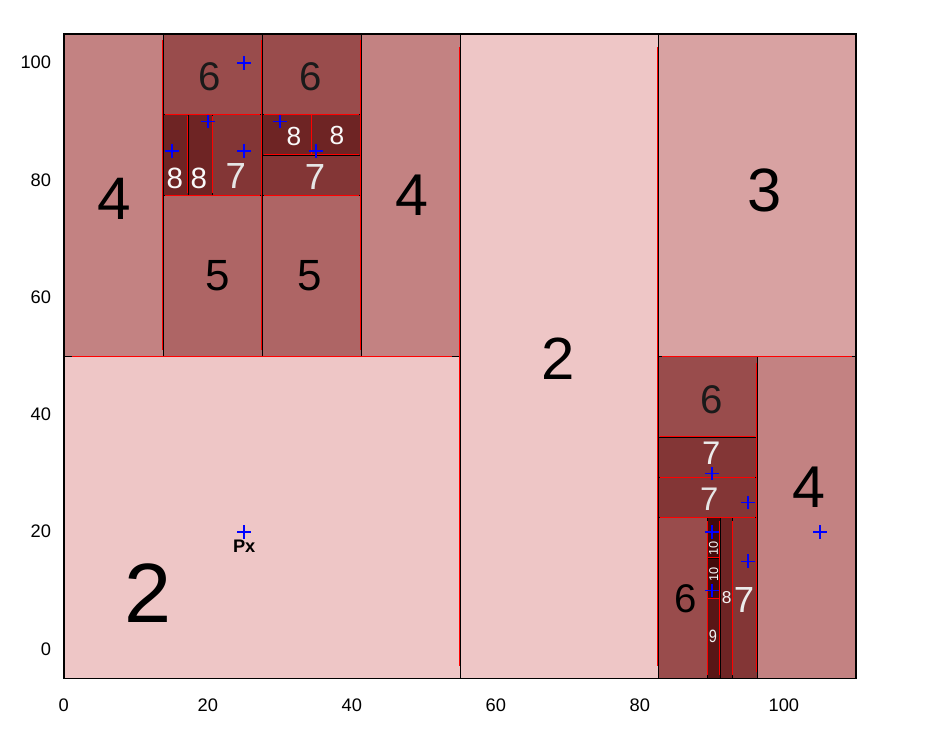}
\caption{Enhanced approach. Rectangles created by recursive splitting.}
\label{fig:example_novelty_gnu}
\end{figure*}

Consider now an example of creating a new HST based on the given input sample
\begin{align*}
    S = \left\{\begin{smallmatrix}
    [25,100],& [30,90], &[20,90],&[35,85],\\
    [25,85],&[15,85],&[105,20],&[95,25], \\
    [95,15], &[90,30],&[90,20],&[90,10]
    \end{smallmatrix}\right\}.
\end{align*}
Figure \ref{fig:example_novelty_gnu} shows $S$ in the finished plane created using an enhanced approach. Observe that the split points (red lines) are always in the middle of the previous observable space, and rectangles now always cover the whole ancestor's area. Numbers represent the final leaves' depth (also seen in Figure \ref{fig:example_novelty_tree_color}).

The tree $T$ is created by starting with the tree $T_0$ and expanding further as described by the recursive step until the ending condition is met.
Figure \ref{fig:example_novelty_tree_color} shows the finished tree $T$, trained on the dataset $S$.

\paragraph{Basis step} 
The trivial step is to create the tree $T_0$ with one root vertex $R$ with the experimentally set initial possibility space  $R= \langle 0,110) \times \langle -5,105)$ and no edges $E_0$, such that
    \begin{align*}
        V_0 &= \{R\},&
        E_0 &= \emptyset,&
        T_0 &= (V_0, E_0).
    \end{align*}

\paragraph{First recursive step}
     Since $R \in L_0$, we create two rectangles $R_l$, $R_r$ by selecting a random dimension $d=0$.
    \begin{align*}
        r_0 &= \langle 0, 110), &
        s &= \frac{0 + 110}{2} = 55, \\
        R_l &= \langle 0, 55) \times \langle -5,105), &
        R_r &= \langle 55, 110) \times \langle -5,105).
    \end{align*}
     New tree $T_1$ is then
    \begin{align*}
    V_1 &= \{R, R_l, R_r\}, &
    E_1 &= \{(R, R_l), (R, R_r)\}, \\
    T_1 &= (V_1, E_1).
    \end{align*}
    Checking for the ending condition, set $L_1$ contains precisely two elements $R_l$ and $R_r$; hence, we continue.
     
\paragraph{Second recursive step}
    Since there are vertices in $L_1$ left to be examined, we continue with recursive steps.
    Since $R_l \in L_1$ (resp. $R_r \in L_1$), we create two new rectangles $R_{ll}$, $R_{lr}$ -- left column (resp. $R_{rl}$, $R_{rr}$ -- right column) by selecting a random dimension $d=1$ (resp. $d=0$).
    \begin{align*}
        r_{l1} &= \langle -5, 105)& r_{r0} &= \langle 55, 110) \\
        s_l &= 50 & s_r&=82.5\\
        R_{ll} &= \langle 0, 55) \times \langle -5,50) & R_{rl} &= \langle 55, 82.5) \times \langle -5,105)\\
        R_{lr} &= \langle 0, 55) \times \langle 50,105) & R_{rr} &= \langle 82.5, 110) \times \langle -5,105)
    \end{align*}
 New tree $T_2$ is then
    \begin{align*}
        V_2 &= \{R, R_l, R_r, R_{ll}, R_{lr}, R_{rl}, R_{rr}\} \\
        E_2 &= \left\{\begin{smallmatrix}
        (R, R_l), &(R, R_r), &(R_l, R_{ll}),\\
        (R_l, R_{lr}),& (R_r, R_{rl}),& (R_r, R_{rr})
        \end{smallmatrix}\right\} \\
        T_2 &= (V_2, E_2)
    \end{align*}
\paragraph{Termination} We recheck the ending condition. The set of leaves $L_2$ now contains two hyperrectangles $R_{lr}$ and $R_{rr}$.
This goes on until the ending condition is met. The resulting tree $T_8$ is depicted in Figure \ref{fig:example_novelty_tree_color}.

\subsection{Evaluating the decision tree}
The evaluation of this decision tree is more straightforward since the examined data point is always contained in the possibility space hyperrectangle of some vertex in each level of depth until a leaf is visited.
The evaluation starts in the root vertex. The initial possibility space should be reasonable enough to cover the whole domain of a given problem. Until the leaf is reached, the examined point recursively visits the descendant within which it is contained.

\begin{example}
\label{ex:regular_point_evaluation_novelty}
    Consider now the evaluation of $a = [105,20]$ on the tree $T$ built in Example \ref{example:novelty_tree_create}.

\begin{enumerate}
    \item  We start with the root $R = \langle 0,110\rangle \times \langle -5, 105 \rangle$.
    The root $R$ has two descendants 
\begin{align*}
    R_l &= \langle 0,55) \times \langle -5, 105),\\
    R_r &= \langle 55,110) \times \langle -5, 105),
\end{align*}
Since $a \in R_r$, we visit $R_r$.
\item Vertex $R_r$ has two descendants
\begin{align*}
    R_{rl} &= \langle 5,82.5) \times \langle -5, 105),\\
    R_{rr} &= \langle 82.5,110) \times \langle -5, 105),
\end{align*}
Since $a \in R_{rr}$, we visit the vertex $R_{rr}$.
\item
We continue recursively for another two steps until the leaf $R_{rrlr}$ is reached. This leaf has a depth of $4$.

\end{enumerate}
   
\end{example}

\begin{figure*}[!t]
\centering
\includegraphics[width=1\textwidth]{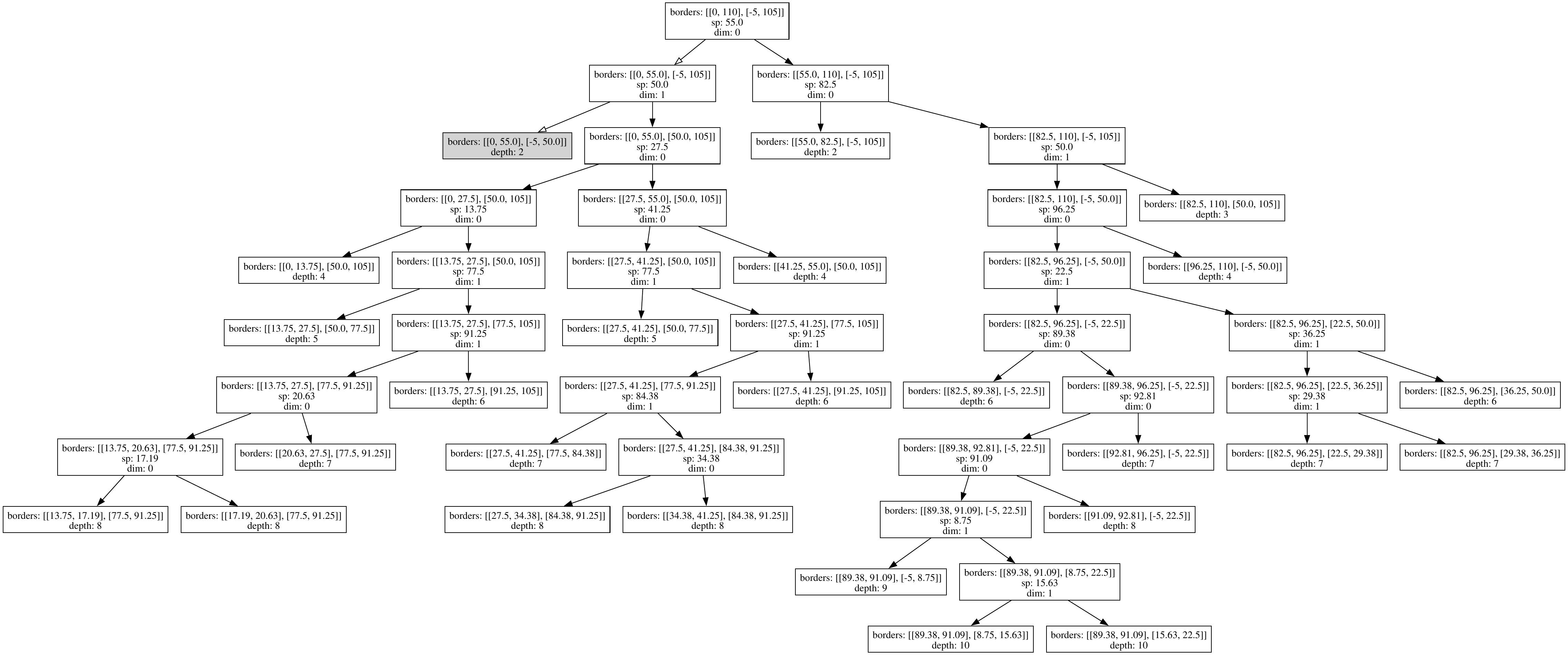}
\caption{Tree constructed using the enhanced novelty approach.}
\label{fig:example_novelty_tree_color}
\end{figure*}

\begin{example}
\label{ex:novelty_point_evaluation_novelty}
    Consider now the evaluation of $a' = [25,20]$ on the tree $T$ built in Example \ref{example:novelty_tree_create}.

\begin{enumerate}
    \item  We start with the root $R = \langle 0,110\rangle \times \langle -5, 105 \rangle$.
    The root $R$ has two descendants 
\begin{align*}
    R_l &= \langle 0,55) \times \langle -5, 105),\\
    R_r &= \langle 55,110) \times \langle -5, 105),
\end{align*}
Since $a' \in R_l$, we visit $R_l$.
\item Vertex $R_l$ has two descendants
\begin{align*}
    &R_{ll} = \langle 0,55) \times \langle -5, 50),&
    &R_{lr} = \langle 0,5) \times \langle 50, 105),
\end{align*}
Since $a' \in R_{ll}$, we visit the vertex $R_{ll}$.
\item
Since $R_{ll}$ is a leaf, we end here, and the reached leaf's depth is $2$. The leaf is marked gray in Figure \ref{fig:example_novelty_tree_color}.
\end{enumerate}
\end{example}
Note that each evaluated point of the given possibility space is always contained in each vertex it visits.
Hence $a \in R_{rrlr} \subset R_{rrl} \subset R_{rr} \subset R_{r} \subset R$ in Example \ref{ex:regular_point_evaluation_novelty}
and $a' \in R_{ll} \subset R_{l} \subset R$ as shown in Example \ref{ex:novelty_point_evaluation_novelty}.


\section{Expected value of depth}
One conclusive method to evaluate both algorithms for subsequent comparison is calculating the expected value of depth. It takes individual points and calculates the probability that a point in a given algorithm will reach that particular depth. It is then possible to compare the expected value of depth of the individual points with each other and see how different they are, giving a scale of abnormality.

For the calculations, we consider the points from the examples above.

\subsection{Original approach}
To show the resulting depth using the original approach, we find all possible paths that would isolate the given point. Then, for each of the possible paths, we calculate its probability.




For the point $[25,85]$, consider one of the possible paths that orphan the given point.

We start with the whole observable space, a range of $\langle 15, 105\rangle$ for the first dimension $x$ and $\langle 10,100\rangle$ for the dimension $y$.
First, the dimension $x$ and a split point in the $\langle 95, 105\rangle$ range were chosen (the split point is a random value from this interval). The possible range where the split point could have been chosen was $\langle 15,105\rangle$.
The probability of selecting $x$ is $0.5$ due to the two-dimensional setting.
The probability of the split point being from the given range is $\frac{105-95}{105-15}$, where the nominator is the size of the favourable range, and the denominator is the size of the whole possible range.
The probability of such event is then $$\frac{1}{2}\cdot\frac{105-95}{105-15} = \frac{1}{18}.$$
Since the given point is not yet orphaned, we continue this way. The observable space is now scaled down due to the split point to the range of $\langle 15, 95\rangle$ for the first dimension $x$ and $\langle 10,100\rangle$ for the second $y$. Table \ref{prob_table_25_85} shows the rest of the probabilities for this path.

\begin{table}[!t]
\centering
\caption{Probabilities of depths for point $[25,85]$.}
\label{prob_table_25_85}
\resizebox{\columnwidth}{!}{%
\begin{tblr}{
    width=\linewidth,
    hspan=minimal,
    cells={font=\footnotesize},
    colspec={c c c c c},
    row{1}={guard},
    column{1-5}={mode=math}
}
Start space & Dim. & Split range & Probability & End space \\
\hline
\langle 15, 105\rangle \times \langle 10, 100\rangle & x & \langle 95, 105\rangle &  \frac{1}{2}\cdot\frac{105-95}{105-15} = \frac{1}{18} & \langle 15, 95\rangle \times \langle 10, 100\rangle \\
\langle 15, 95\rangle \times \langle 10, 100\rangle & x & \langle 90, 95\rangle &  \frac{1}{2}\cdot\frac{95-90}{95-15} = \frac{1}{32} & \langle 15, 90\rangle \times \langle 10, 100\rangle \\
\langle 15, 90\rangle \times \langle 10, 100\rangle & y & \langle 85, 90\rangle &  \frac{1}{2}\cdot\frac{90-85}{100-10} = \frac{1}{36} & \langle 15, 90\rangle \times \langle 10, 85\rangle \\
\langle 15, 90\rangle \times \langle 10, 85\rangle & y & \langle 10, 20\rangle &  \frac{1}{2}\cdot\frac{20-10}{85-10} = \frac{1}{15} & \langle 15, 90\rangle \times \langle 20, 85\rangle \\
\langle 15, 90\rangle \times \langle 20, 85\rangle & x & \langle 35, 90\rangle &  \frac{1}{2}\cdot\frac{90-35}{90-15} = \frac{11}{30} & \langle 15, 35\rangle \times \langle 85, 85\rangle \\
\langle 15, 35\rangle \times \langle 85, 85\rangle & x & \langle 25, 35\rangle &  \phantom{\frac{1}{2}\cdot}\frac{35-25}{35-15} = \frac{1}{2} & \langle 15, 25\rangle \times \langle 85, 85\rangle \\
\langle 15, 25\rangle \times \langle 85, 85\rangle & x & \langle 15, 25\rangle &  \phantom{\frac{1}{2}\cdot}\frac{25-15}{25-15} = \frac{1}{1} & \langle 25, 25\rangle \times \langle 85, 85\rangle
\end{tblr}
}
\end{table}


Note that rows $6$ and $7$ no longer contain the probability of selecting the dimensions since the second dimension cannot be chosen as it would not isolate any point (see the startspace's $y$ as $\langle 85, 85\rangle$).
The evaluation ends after seven splits (depth = $7$) since that is the last split to isolate the given point.
The probability of this case is then $\frac{1}{18}\cdot\frac{1}{32}\cdot\dots\cdot\frac{1}{2}\cdot 1$.

If we do this for all possible cases, we get the probabilities for the individual depths that could result in orphaning the given point. The first row in Table \ref{table_big_original} shows the values for all possible depths for the point $[25,100]$. 
The rest of the rows of this table show the probabilities of depth for the remaining points.
This is later used to calculate the expected value of depth.
Due to the symmetricity, the probabilities for the points in the bottom right corner in Figure \ref{fig:example_noutlier_gnu} are the same as those in the top left corner, so we only show the latter.

Let us recall the initial problem in Section \ref{sec:revisiting}.
If we consider the novelty point $[25,20]$, the evaluation results in the same path as the point $[25,85]$ in Table \ref{prob_table_25_85}. However, the novelty point no longer fits in the start space for the sixth and seventh rows. Nevertheless, because the point $[25,20]$ satisfies $25 \le z \in \langle 25, 35 \rangle$,  $[25,20]$ is assigned to the vertex $\langle 15, 25 \rangle \times \langle 85, 85 \rangle$ even though $[25,20] \notin \langle 15, 25 \rangle \times \langle 85, 85 \rangle$.

Note that the path for the novelty point $[25,20]$ is a path of the $[25,85]$ (the vertical neighbour in Figure \ref{fig:example_noutlier_gnu}) or the $[90,20]$ (the horizontal neighbour).

\begin{sidewaystable}[!t]
\caption{Probabilities for individual data points, original approach.}
\label{table_big_original}
\begin{tblr}{
    width=\linewidth,
    hspan=minimal,
    cells={font=\footnotesize},
    cell{1}{1-11}={halign=c},
    column{odd}={gray9},
    colspec={
    c |
    S[round-mode=places ,round-precision=2, output-exponent-marker=E, table-format=1.2e+1]
    S[round-mode=places ,round-precision=2, output-exponent-marker=E, table-format=1.2e+1]
    S[round-mode=places ,round-precision=2, output-exponent-marker=E, table-format=1.2e+1]
    S[round-mode=places ,round-precision=2, output-exponent-marker=E, table-format=1.2e+1]
    S[round-mode=places ,round-precision=2, output-exponent-marker=E, table-format=1.2e+1]
    S[round-mode=places ,round-precision=2, output-exponent-marker=E, table-format=1.2e+1]
    S[round-mode=places ,round-precision=2, output-exponent-marker=E, table-format=1.2e+1]
    S[round-mode=places ,round-precision=2, output-exponent-marker=E, table-format=1.2e+1]S[round-mode=places ,round-precision=2, output-exponent-marker=E, table-format=1.2e+1]
    S[round-mode=places ,round-precision=2, output-exponent-marker=E, table-format=1.2e+1]
    S[round-mode=places ,round-precision=2, output-exponent-marker=E, table-format=1.2e+1]
    S[round-mode=places ,round-precision=2, output-exponent-marker=E, table-format=1.2e+1]
    },
    row{1}={guard},
    column{1}={guard, mode=math}
}
 \diagbox{Point}{Depth} & 1 & 2 & 3 & 4 & 5 & 6 & 7 & 8 & 9 & 10 & 11 \\
 \hline
\left[25, 100\right] & 5.5555555556E-02 & 2.5103323737E-01 & 2.5570744502E-01 & 2.4773022596E-01 & 1.3819675888E-01 & 4.3771373491E-02 & 7.2084454376E-03 & 7.4839864762E-04 & 4.6942907037E-05 & 1.5970541667E-06 & 1.9690558403E-08\\
\left[20, 90\right] & 0 & 3.5539215686E-02 & 1.7775615069E-01 & 4.0676493336E-01 & 2.6984418116E-01 & 9.2561204648E-02 & 1.5747066825E-02 & 1.6759728793E-03 & 1.0748807222E-04 & 3.7394349905E-06 & 4.7254206811E-08\\
\left[30, 90\right] & 0 & 1.3368055556E-02 & 1.6859439869E-01 & 4.1552713020E-01 & 2.8292729344E-01 & 9.9767533944E-02 & 1.7705833609E-02 & 1.9721214421E-03 & 1.3271521257E-04 & 4.8533703768E-06 & 6.4529234241E-08\\
\left[35, 85\right] & 0 & 9.7486772487E-02 & 2.5792650002E-01 & 3.2462216793E-01 & 2.4112764708E-01 & 6.6916307461E-02 & 1.0752400748E-02 & 1.0967974134E-03 & 6.8988815370E-05 & 2.3864551481E-06 & 3.1601016673E-08 \\
\left[25, 85\right] & 0 & 0 & 2.4722562636E-02 & 3.0688106140E-01 & 4.7133742736E-01 & 1.6542794157E-01 & 2.8432586746E-02 & 3.0013428173E-03 & 1.9044983168E-04 & 6.5444606635E-06 & 8.3183446589E-08 \\
\left[15, 85\right] & 2.7777777778E-02 & 1.2482638889E-01 & 2.5516544084E-01 & 3.0935396980E-01 & 2.1892437075E-01 & 5.5415608206E-02 & 7.8217752227E-03 & 6.7869855413E-04 & 3.5013796136E-05 & 9.4653744554E-07 & 9.6253761386E-09\\
\hline
\left[20, 25\right] & 0 & 3.5693536674E-02 & 1.8350184759E-01 & 3.4881778276E-01 & 3.4168298237E-01 & 8.0460443076E-02 & 9.2155826505E-03 & 6.0623378611E-04 & 2.1232873030E-05 & 3.5512746939E-07 & 3.0996586909E-09
\end{tblr}

\end{sidewaystable}

\subsection{Novelty approach}
To provide an argument for the resulting depth using the novelty approach, we calculate the expected value of depth for each point $p$, considering all possible trees that would isolate $p$ and their respective paths. Contrary to the original approach, we can get potentially infinite splits, resulting in variable depth.
We start with given range $\langle 0, 110.0\rangle \times \langle -5.0, 105.0\rangle$.

\paragraph{The first point to consider is \([25,100]\).} There is only one way to isolate this point: using three horizontal splits (H). This is the only scenario S1 for this point. Table \ref{table_25_100} shows the probabilities for individual depths. Since the vertical splits (V) do not contribute to the isolation of a given point, there could be any number of them. Hence, we get the expected value of depth as
$$\sum_{n=3}^{\infty}\binom{n-1}{2}\cdot \frac{1}{2^n}\cdot n = 6.$$

\begin{table}[!t]
\centering
\caption{Probabilities of depths for point $[25,100]$.}
\label{table_25_100}
\resizebox{\columnwidth}{!}{%
\begin{tblr}{
    width=\linewidth,
    hspan=minimal,
    cells={font=\footnotesize},
    colspec={c| c | c | c | c},
    column{odd}={gray9},
    row{1}={guard},
    column{1-5}={guard, mode=math}
}
 \diagbox{Depth}{Probab.} & V & H & S1 & \sum \\
 \hline
3 & 0 & 3 & \binom{2}{2}\cdot\frac{1}{2^3} & \frac{1}{8} \\
4 & 1 & 3 & \binom{3}{2}\cdot\frac{1}{2^4} & \frac{3}{16} \\
5 & 2 & 3 & \binom{4}{2}\cdot\frac{1}{2^5} & \frac{3}{16} \\
\vdots & \vdots & \vdots & \vdots & \vdots  \\
k & k-3 & 3 & \binom{k-1}{2}\cdot \frac{1}{2^k} & \binom{k-1}{2}\cdot \frac{1}{2^k} \\
\vdots & \vdots & \vdots & \vdots & \vdots \\
\hline
\sum & - & - & 1 & 1
\end{tblr}
}
\end{table}

\paragraph{The second point in our sample is $[20,90]$.} Now, there are more ways to isolate this point. That is, by exactly four vertical splits and up to four horizontal splits (S1) or at least five horizontal splits along with two or three vertical splits (S2X).
To get the expected value for all the possible scenarios, we must calculate their probabilities. Table \ref{table_20_90} shows the probabilities for individual depths.

To simplify the second scenario --- two vertical and five horizontal splits --- we divide it into two sub-scenarios.
\begin{enumerate}
    \item Sub-scenario (S2V), where the last split is vertical (exactly two vertical splits).
    \item Sub-scenario (S2H), where the last split is horizontal (exactly five horizontal splits).
\end{enumerate}

The expected value of depth for the point $[20,90]$ is
\begin{multline*}
  \sum_{n=4}^{8}\binom{n-1}{3}\cdot \frac{1}{2^n}\cdot n + \sum_{n=7}^{\infty}\binom{n-1}{1}\cdot \frac{1}{2^n}\cdot n +\\ +\sum_{n=7}^{8}\binom{n-1}{4}\cdot \frac{1}{2^n}\cdot n \doteq 6.82.
\end{multline*}

\begin{table}[!t]
\caption{Probabilities of depths for point $[20,90]$.}
\label{table_20_90}
\centering
\resizebox{\columnwidth}{!}{%
\begin{tblr}{
    width=\linewidth,
    hspan=minimal,
    cells={font=\footnotesize},
    colspec={c| c c c | c},
    column{odd}={gray9},
    row{1}={guard},
    column{1-5}={guard, mode=math}
}
 \diagbox{Depth}{Probab.} & S1 & S2V & S2H & \sum \\
 \hline
4 & \binom{3}{3}\cdot \frac{1}{2^4} & 0 & 0 & \frac{1}{16} \\
5 & \binom{4}{3}\cdot\frac{1}{2^5}  &  0 & 0 & \frac{1}{8}\\
6 & \binom{5}{3}\cdot\frac{1}{2^6}  &  0 & 0& \frac{5}{32}\\
7 & \binom{6}{3}\cdot\frac{1}{2^7}  & \binom{6}{1}\cdot\frac{1}{2^7} & \binom{6}{4}\cdot\frac{1}{2^7} & \frac{41}{128} \\
8 & \binom{7}{3}\cdot\frac{1}{2^8}  & \binom{7}{1}\cdot\frac{1}{2^8} & \binom{7}{4}\cdot\frac{1}{2^8} & \frac{77}{256}\\
9 & 0 & \binom{8}{1}\cdot\frac{1}{2^9} & 0 & \frac{1}{64}\\
\vdots & \vdots & \vdots & \vdots & \vdots\\
k & 0 & \binom{k-1}{1}\cdot \frac{1}{2^k} & 0 & (k-1)\cdot\frac{1}{2^k}\\
\vdots & \vdots & \vdots & \vdots & \vdots \\
\hline
\sum & \frac{163}{256} & \frac{7}{64} & \frac{65}{256} & 1
\end{tblr}
}
\end{table}

\paragraph{The third point in the sample is $[15,85]$.} We need four vertical splits or two or three vertical and five horizontal splits to isolate this point. That is the same scenario as the previous point $[20,90]$; hence, we get the same depths, resulting in the same probabilities.

\paragraph{The fourth point in the sample is $[30,90]$.} We need either exactly five vertical splits (S1) or two, three or four vertical and exactly five horizontal splits to isolate this point. That gives, again, two sub-scenarios. First, the last split is vertical (S2V), and conversely, the last is horizontal (S2H).
Table \ref{table_30_90} shows the probabilities for individual depths.


The expected value of depth for the point $[30,90]$ is

\begin{multline*}
\sum_{n=5}^{9}\binom{n-1}{4}\cdot \frac{1}{2^n}\cdot n + \sum_{n=7}^{\infty}\binom{n-1}{1}\cdot \frac{1}{2^n}\cdot n+\\
+ \sum_{n=7}^{9}\binom{n-1}{4}\cdot \frac{1}{2^n}\cdot n \doteq 7.82
\end{multline*}

\begin{table}[!t]
\caption{Probabilities of depths for point $[30,90]$.}
\label{table_30_90}
\centering
\resizebox{\columnwidth}{!}{%
\begin{tblr}{
    width=\linewidth,
    hspan=minimal,
    cells={font=\footnotesize},
    colspec={c | c c c | c },
    column{odd}={gray9},
    row{1}={guard},
    column{1-5}={guard, mode=math}
}
 \diagbox{Depth}{Probab.} & S1 & S21 & S22 & \sum  \\
 \hline
5 & \binom{4}{4}\cdot\frac{1}{2^5} & 0 & 0 & \frac{1}{32}\\
6 & \binom{5}{4}\cdot\frac{1}{2^6} & 0 & 0 & \frac{5}{64}\\
7 & \binom{6}{4}\cdot\frac{1}{2^7} & \binom{6}{1}\cdot\frac{1}{2^7} & \binom{6}{4}\cdot\frac{1}{2^7} & \frac{9}{32}\\
8 & \binom{7}{4}\cdot\frac{1}{2^8} & \binom{7}{1}\cdot\frac{1}{2^8} & \binom{7}{4}\cdot\frac{1}{2^8} & \frac{77}{256}\\
9 & \binom{8}{4}\cdot\frac{1}{2^9} & \binom{8}{1}\cdot\frac{1}{2^9} & \binom{8}{4}\cdot\frac{1}{2^9} & \frac{37}{128}\\
10 & 0 & \binom{9}{1}\cdot\frac{1}{2^{10}} & 0 & \frac{9}{1024}\\
\vdots & \vdots & \vdots & \vdots & \vdots \\
k & 0 & \binom{k-1}{1}\cdot \frac{1}{2^k} & 0 & (k-1)\cdot \frac{1}{2^k} \\
\vdots & \vdots & \vdots & \vdots & \vdots \\
\hline
\sum & \frac{1}{2} & \frac{7}{64} & \frac{25}{64} & 1
\end{tblr}
}
\end{table}

\paragraph{The fifth point in the sample is $[35,85]$.}  We need four vertical splits to isolate this point. That is, any number of horizontal splits and four vertical splits are mixed so that the last split is always vertical.
We get the expected value of depth as follows
$$\sum_{n=4}^{\infty}\binom{n-1}{3}\cdot \frac{1}{2^n}\cdot n = 8.$$

Table \ref{table_35_85} shows the probabilities for individual depths for the fifth point.

\begin{table}[!t]
\caption{Probabilities of depths for point $[35,85]$.}
\label{table_35_85}
\centering
\resizebox{\columnwidth}{!}{%
\begin{tblr}{
    width=\linewidth,
    hspan=minimal,
    cells={font=\footnotesize},
    colspec={c | c | c},
    column{odd}={gray9},
    row{1}={guard},
    column{1-3}={guard, mode=math}
}
 \diagbox{Depth}{Probab.} & S1 & \sum \\
 \hline
4 & \binom{3}{3}\cdot\frac{1}{2^4} & \frac{1}{16} \\
5 & \binom{4}{3}\cdot\frac{1}{2^5} &  \frac{1}{8}\\
6 & \binom{5}{3}\cdot\frac{1}{2^6} & \frac{5}{32} \\
\vdots & \vdots & \vdots \\
k & \binom{k-1}{3}\cdot \frac{1}{2^k} & \binom{k-1}{3}\cdot \frac{1}{2^k}\\
\vdots & \vdots & \vdots \\
\hline
\sum & 1 & 1
\end{tblr}
}
\end{table}

\paragraph{The sixth point in the sample is the point  $[25,85]$.} To isolate this point, we need five vertical and three or four horizontal splits (S1V and S1H) or four vertical and at least five horizontal splits (S2V and S2H).
Table \ref{table_25_85} shows the probabilities for individual depths for the above scenarios.




The expected value of depth for the point $[25,85]$ is:
\begin{multline*}
\sum_{n=8}^{9}\binom{n-1}{4}\cdot \frac{1}{2^n}\cdot n
+ \sum_{n=8}^{\infty}\binom{n-1}{2}\cdot \frac{1}{2^n}\cdot n+\\
+ \sum_{n=9}^{\infty}\binom{n-1}{3}\cdot \frac{1}{2^n}\cdot n 
+ \binom{8}{4}\cdot \frac{1}{2^9} \doteq 9.734
\end{multline*}


\begin{table}[!t]
\caption{Probabilities of depths for point $[25,85]$.}
\label{table_25_85}
\renewcommand{\arraystretch}{1.3}
\centering
\resizebox{\columnwidth}{!}{%
\begin{tblr}{
    width=\linewidth,
    hspan=minimal,
    cells={font=\footnotesize},
    colspec={c | c c c c | c},
    column{odd}={gray9},
    row{1}={guard},
    column{1-6}={guard, mode=math}
}
 \diagbox{Depth}{Probab.} & S1V & S1H & S2V & S2H & \sum \\
 \hline
8 & \binom{7}{4}\cdot\frac{1}{2^8} &  \binom{7}{2}\cdot\frac{1}{2^8} & 0 & 0 & \frac{7}{32}\\
9 & \binom{8}{4}\cdot\frac{1}{2^9} & \binom{8}{2}\cdot\frac{1}{2^9} & \binom{8}{3}\cdot\frac{1}{2^9} & \binom{8}{4}\cdot\frac{1}{2^9} & \frac{7}{16} \\
10 & 0 & \binom{9}{2}\cdot\frac{1}{2^{10}} & \binom{9}{3}\cdot\frac{1}{2^{10}} & 0 & \frac{15}{128}\\
\vdots & \vdots & \vdots & \vdots & \vdots & \vdots \\
k & 0 & \binom{k-1}{2}\cdot\frac{1}{2^k} & \binom{k-1}{3}\cdot \frac{1}{2^k} & 0 & \binom{k}{3}\cdot \frac{1}{2^k}  \\
\vdots & \vdots & \vdots & \vdots & \vdots & \vdots \\
\hline
\sum & \frac{35}{128} & \frac{29}{128} & \frac{93}{256} & \frac{35}{256} & 1
\end{tblr}
}
\end{table}

We can determine that our novelty point $[25,20]$ gets the value of the expected depth of $3$. Table \ref{table_novelty} shows the probabilities for the individual depths. When orphaning this point, we only have two possible sub-scenarios.

If the last split is horizontal, we get
$$\sum_{n=2}^{\infty}\binom{n-1}{0}\frac{1}{2^{n}}\cdot n = 1.5.$$
Conversely, if the last split is vertical, we get
$$\sum_{n=2}^{\infty}\binom{n-1}{0}\frac{1}{2^{n}}\cdot n = 1.5.$$
This can be simplified as
$$\sum_{n=2}^{\infty}\frac{1}{2^{n-1}}\cdot n = 3.$$
The expected value of depth of $[25,85]$ is $3$.

\begin{table}[!t]
\centering
\caption{Probabilities of depths for the novelty point $[25,20]$.}
\label{table_novelty}
\begin{tblr}{
    width=\linewidth,
    hspan=minimal,
    cells={font=\footnotesize},
    colspec={c| c c |c},
    column{odd}={gray9},
    row{1}={guard},
    column{2-5}={guard, mode=math}
}
 \diagbox{Depth}{Probab.} & S1V & S1H & \sum \\
 \hline
2 & \binom{1}{0}\cdot\frac{1}{2^2} &  \binom{1}{0}\cdot\frac{1}{2^2} & \frac{1}{2}\\
3 & \binom{2}{0}\cdot\frac{1}{2^3} & \binom{2}{0}\cdot\frac{1}{2^3} &  \frac{1}{4}\\
4 & \binom{3}{0}\cdot\frac{1}{2^4} & \binom{3}{0}\cdot\frac{1}{2^4} & \frac{1}{8}\\
\vdots & \vdots & \vdots &\vdots \\
k & \binom{k-1}{0}\cdot\frac{1}{2^k} & \binom{k-1}{0}\cdot\frac{1}{2^k}&\frac{1}{2^{k-1}} \\
\vdots & \vdots & \vdots & \vdots\\
\hline
\sum & \frac{1}{2} & \frac{1}{2} & 1 \\
\end{tblr}
\end{table}

Table \ref{table_big_novelty} shows the aggregated sums for individual depths for comparison with the expected values for the original approach in Table \ref{table_big_original}.

\begin{sidewaystable}[!t]
\caption{Probabilities for individual data points, enhanced approach.}
\label{table_big_novelty}
\begin{tblr}{
    width=\linewidth,
    hspan=minimal,
    cells={font=\footnotesize},
    cell{1}{1-11}={halign=c},
    column{odd}={gray9},
    colspec={
    c |
    S[round-mode=places ,round-precision=2, output-exponent-marker=E, table-format=1.2e+1]
    S[round-mode=places ,round-precision=2, output-exponent-marker=E, table-format=1.2e+1]
    S[round-mode=places ,round-precision=2, output-exponent-marker=E, table-format=1.2e+1]
    S[round-mode=places ,round-precision=2, output-exponent-marker=E, table-format=1.2e+1]
    S[round-mode=places ,round-precision=2, output-exponent-marker=E, table-format=1.2e+1]
    S[round-mode=places ,round-precision=2, output-exponent-marker=E, table-format=1.2e+1]
    S[round-mode=places ,round-precision=2, output-exponent-marker=E, table-format=1.2e+1]
    S[round-mode=places ,round-precision=2, output-exponent-marker=E, table-format=1.2e+1]
    S[round-mode=places ,round-precision=2, output-exponent-marker=E, table-format=1.2e+1]
    S[round-mode=places ,round-precision=2, output-exponent-marker=E, table-format=1.2e+1]
    S[round-mode=places ,round-precision=2, output-exponent-marker=E, table-format=1.2e+1]
    },
    row{1}={guard},
    column{2-13}={mode=math},
    column{1}={guard, mode=math}
}
 \diagbox{Point}{Depth} & 1 & 2 & 3 & 4 & 5 & 6 & 7 & 8 & 9 & 10 & >10 \\
 \hline
\left[25, 100\right] & 0 & 0 & 1.250e-1 & 1.880e-1 & 1.880e-1 & 1.560e-1 & 1.170e-1 & 8.200e-2 & 5.470e-2 & 3.520e-2 & 4.07e-02 \\
\left[20, 90\right] & 0 & 0 & 0 & 6.250e-2 & 1.250e-1 & 1.560e-1 & 3.200e-1 & 3.010e-1 & 1.560e-2 & 8.790e-3 & 6.23e-03 \\
\left[30, 90\right] & 0 & 0 & 0 & 0 & 3.130e-2 & 7.810e-2 & 2.810e-1 & 3.010e-1 & 2.890e-1 & 8.790e-3 & 5.93e-03 \\
\left[35, 85\right] & 0 & 0 & 0 & 6.250e-2 & 1.250e-1 & 1.560e-1 & 1.560e-1 & 1.370e-1 & 1.090e-1 & 8.200e-2 & 1.14e-01 \\
\left[25, 85\right] & 0 & 0 & 0 & 0 & 0 & 0 & 0 & 2.190e-1 & 4.380e-1 & 1.170e-1 & 1.45e-01 \\
\left[15, 85\right] & 0 & 0 & 0 & 6.250e-2 & 1.250e-1 & 1.560e-1 & 3.200e-1 & 3.010e-1 & 1.560e-2 & 8.790e-3 & 6.23e-03\\
\hline
\left[20, 25\right] & 0 & 5.000e-1 & 2.500e-1 & 1.250e-1 & 6.250e-2 & 3.130e-2 & 1.560e-2 & 7.810e-3 & 3.910e-3 & 1.950e-3 & 9.53e-04
\end{tblr}
\end{sidewaystable}

\subsection{Conclusion}
The expected values of depths (EXD) for the HST algorithm are shown in the section above.
To compare the values with the expected values of depths for the original isolation forest, we sum the rows in Table \ref{table_big_original} multiplied by respective depth values. Table \ref{table_ex_comparison} shows the side-by-side comparison. The important outcome is the ratio of expected depths in the respective column. As seen in this table, the difference in depth for the novelty point in the case of the original approach is relatively small compared to the difference in the HST forest.
This example proved that it is feasible for the HST algorithm to encapsulate the previously unseen data points in the higher leaves of the tree, making novelty detection possible.
This example closely resembles other similar novelty detection problems.

\begin{table}[!t]
\caption{Expected values of depths for both algorithms.}
\label{table_ex_comparison}
\centering
\begin{tblr}{
    width=\linewidth,
    cells={font=\footnotesize},
    colspec={c | 
    S[table-format=1.3, round-mode=places ,round-precision=3] 
    S[table-format=1.3]},
    column{1-3}={mode=math},
    column{1}={preto=[, appto=]},    
    row{1}={guard,mode=text}
}
point & EXD\ outlier & EXD\ novelty \\
\hline
25,100 & 3.32616229 & 6\\
20,90 & 4.26063731 & 6.82\\
30,90 & 4.34883099 & 7.82\\
35,85& 3.95906409 & 8\\
25,85 & 4.87576598 & 9.734\\
15,85 & 3.76796497 & 6.82\\
\hline
20,25 & 4.27789495 & 3
\end{tblr}
\end{table}

\section{Discussion and Conclusions}
\label{sec:conclusion}
In Section \ref{sec:novelty_isolation_forest} we presented the initial problem of novelty detection and highlighted the exact areas where the original Isolation Forest can be enhanced to be successfully used in novelty detection scenarios.
Later in this section, we proposed our solution based on the Half-Space Tree algorithm.
Through the examples in Section \ref{example:novelty_tree_create}, we show how to build HST using a descriptive dataset.
Experiments in this paper fully demonstrate that, using our proposed enhancement, it is possible to successfully isolate previously unseen novelty data points, with their depth being reasonably different from the regular trained-on data points.
This assumption is later solidified with proofs that show the theoretical outcomes if all possible solutions were assessed. Namely, Table \ref{table_big_novelty} shows that the expected depth of the novelty point when evaluated using the enhanced approach differs much more significantly from other points as opposed to the original approach seen in Table \ref{table_big_original}.

In summary, the research presented herein introduces a novel approach that effectively handles specific novelty points, as evidenced through a detailed example.
Notably, comparisons between the depths of the Isolation Forest algorithm and the Half-Space Tree algorithm indicate significant improvement in detecting novelty points, suggesting enhanced performance in these areas.
Such findings affirm the practical utility of the new approach and highlight its potential adaptability to various situations within the algorithmic framework. 
However, using this approach can no longer detect outliers, as it now works in semi-supervised mode. 
Also, there is a need to choose the proper starting range and other hyperparameters. 
This is why future work will focus on establishing optimal range settings at the onset, addressing unique scenarios within the algorithmic structure, and validating these enhancements through comprehensive benchmarks.






\end{document}